\documentclass{article}
\usepackage{spconf,amsmath,graphicx,hyperref}
\usepackage{booktabs}
\usepackage{amssymb}
\usepackage{wrapfig}
\usepackage{tablefootnote}

\title{DreamVAR: Taming Reinforced Visual Autoregressive Model for High-Fidelity Subject-Driven Image Generation}
\name{Xin Jiang\textsuperscript{1}\sthanks{This work was performed at Hidream.ai Inc.}~Jingwen Chen\textsuperscript{3}~Yehao Li\textsuperscript{3}~Yingwei Pan\textsuperscript{3}~Kezhou Chen\textsuperscript{2}~Zechao Li\textsuperscript{1}\sthanks{Corresponding Author: zechao.li@njust.edu.cn}~Ting Yao\textsuperscript{3}~Tao Mei\textsuperscript{3}}

\address{\textsuperscript{1} Nanjing University of Science and Technology ~\textsuperscript{2} University of Science and Technology of China\\\textsuperscript{3} HiDream.ai Inc.}
\begin{document}
\maketitle
\begin{abstract}
Recent advances in subject-driven image generation using diffusion models have attracted considerable attention for their remarkable capabilities in producing high-quality images.
Nevertheless, the potential of Visual Autoregressive (VAR) models, despite their unified architecture and efficient inference, remains underexplored.
In this work, we present DreamVAR, a novel framework for subject-driven image synthesis built upon a VAR model that employs next-scale prediction.
Technically, multi-scale features of the reference subject are first extracted by a visual tokenizer.
Instead of interleaving these conditional features with target image tokens across scales, our DreamVAR pre-fills the full subject feature sequence prior to predicting target image tokens.
This design simplifies autoregressive dependencies and mitigates the train-test discrepancy in multi-scale conditioning scenario within the VAR paradigm.
DreamVAR further incorporates reinforcement learning to jointly enhance semantic alignment and subject consistency.
Extensive experiments demonstrate that DreamVAR achieves superior appearance preservation compared to leading diffusion-based methods.
\end{abstract}
\section{Introduction}
Subject-driven image generation constitutes a fundamental task in visual generation, requiring models to follow textual prompts while faithfully preserving the visual appearance of reference subjects.
While numerous efforts~\cite{ruiz2023dreambooth,ipa,tan2024ominicontrol,wu2025less,wanvton,gao2025styleshot,chen2023controlstyle} have been dedicated to achieving high-quality subject customization using diffusion models, the potential of Visual Autoregressive (VAR) models for such fine-grained controllability remains largely underexplored.
Accordingly, this work focuses on VAR models for subject-driven synthesis.

Generally, VAR models~\cite{han2025infinity,tian2024visual,mao2025varedit} define autoregressive learning in image generation as next-scale prediction or next-resolution prediction.
Compared to raster-scan autoregressive (AR) models~\cite{ramesh2021dalle}, masked AR models~\cite{yao2025denoising,zheng2025hierarchical} and diffusion models~\cite{ho2020ddpm,rombach2022high,cai2025hidream}, VAR models achieve faster inference speed.
%
Moreover, the unified architecture of VAR models enables flexible integration of additional modalities.
Recent efforts have aimed at improving the controllability of VAR models.
%
ControlVAR~\cite{li2024controlvar} extends a VAR model to condition image generation on structural signals such as edges or depth maps.
Furthermore, ARbooth~\cite{chung2025fine} tackles customized image generation by fine-tuning selective layers and subject-specific text embeddings within a VAR model.
Although effective, this approach necessitates per-subject optimization, incurring significant computational overhead for practical deployment.

\begin{figure*}[htbp]
    \centering
    \includegraphics[width=0.88\textwidth]{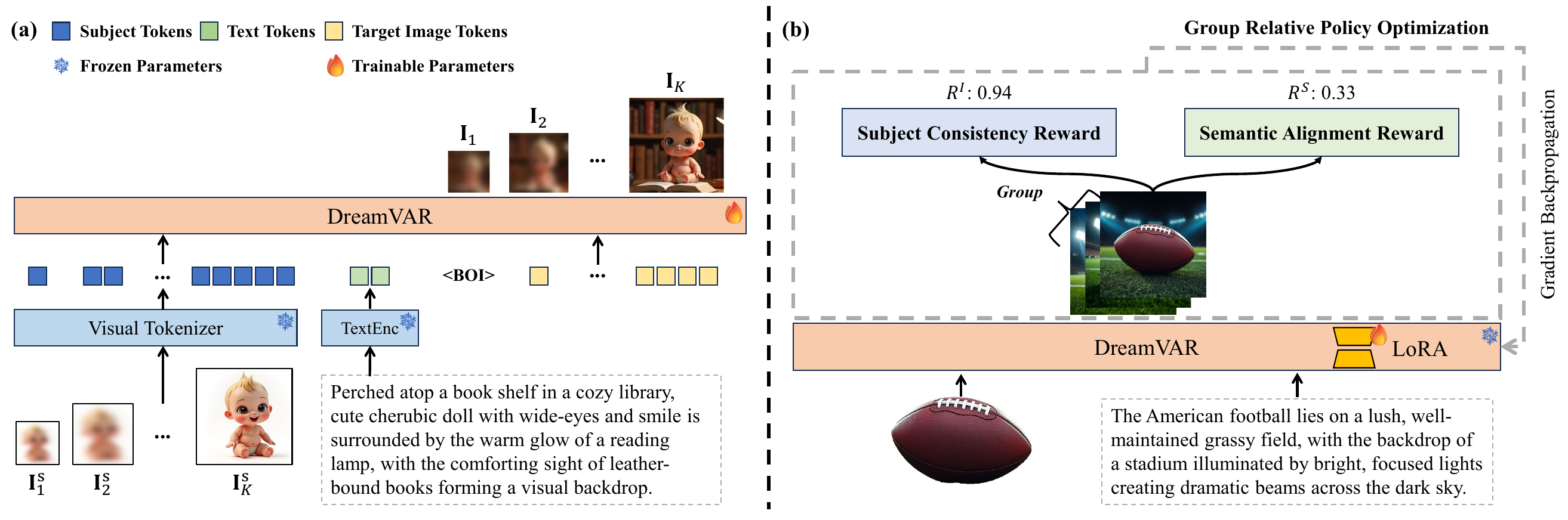}
    \vspace{-4mm}
    \caption{(a) The framework of our DreamVAR for subject-driven image generation, which alleviates the train-test discrepancy through subject feature pre-filling. (b) The pipeline of reinforcement learning with GRPO in our DreamVAR. }
    \label{fig:var}
    \vspace{-4mm}
\end{figure*}

To overcome the above limitations and leverage the merits of VAR models, we present a novel framework named DreamVAR, as shown in Fig.~\ref{fig:var} (a). 
Within the next-scale prediction paradigm, a straightforward way~\cite{li2024controlvar} to inject reference conditions is to concatenate them with the target image tokens at their corresponding scales.
However, we found that this trivial strategy led to suboptimal results.
We argue that this is due to the train-test discrepancy: model predictions are conditioned on ground-truth history during training, but on generated image tokens in inference.
This impact would be further amplified by interleaving unchanged conditional tokens with target image tokens across scales.
To mitigate this discrepancy, our DreamVAR pre-filles the multi-scale features of the reference subject extracted by a visual tokenizer before initiating autoregressive generation.
%
%
Inspired by the success of leveraging reinforcement learning in LLMs~\cite{grpo,openai2024gpt4technicalreport}, we incorporate Group Relative Policy Optimization (GRPO)~\cite{grpo} into our DreamVAR with multiple rewards for subject consistency and semantic alignment (shown in Fig.~\ref{fig:var} (b)), which finally improves generation quality.
Extensive experiments show that our DreamVAR achieves strong subject consistency, surpassing leading diffusion-based models.
\section{Methodology}
\subsection{Preliminary: Next-Scale Prediction}
Visual Autoregressive~(VAR) models~\cite{han2025infinity,tian2024visual} reformulate the autoregressive process from next-token prediction to next-scale prediction.
Formally, let $\mathbf{I} = (\mathbf{I}_{1}, \mathbf{I}_{2}, ..., \mathbf{I}_{K})$ denote token maps of $K$ different resolutions or scales for an image, the autoregressive likelihood of VAR models is defined as:
\begin{align}
    p(\mathbf{I}_{1}, \mathbf{I}_{2}, ..., \mathbf{I}_{K}) = \prod_{k=1}^{K} p(\mathbf{I}_{k} | \mathbf{I}_{<k}, \mathbf{C}),
\label{eq:var}
\end{align}
where $\mathbf{C}$ denotes extra conditions for image generation (e.g., text prompt).
Practically, the 2D spatial token map $\mathbf{I}_{k}$ is usually flattened into a 1D token sequence for computation.

\subsection{DreamVAR}
Our proposed DreamVAR for subject-driven image generation builds upon a pre-trained text-to-image VAR model (Infinity~\cite{han2025infinity}), which employs multi-scale visual tokenizer and bitwise self-correction to improve generation quality. Please refer to~\cite{han2025infinity} for more details about Infinity. 

As described previously in the introduction, VAR models tend to suffer from the train-test discrepancy due to teacher-forcing training schema.
This impact would be amplified in multi-scale conditional generation when trivially interleaving conditional tokens with target image tokens, leading to degraded performance.
To overcome this, we propose pre-filling all the conditional tokens of the reference subject prior to the autoregressive prediction of target image tokens. The overall framework is illustrated in Fig.\ref{fig:var} (a).

Specificially, our DreamVAR extracts multi-scale features $\mathbf{I}^{s} = (\mathbf{I}_{1}^{s}, \mathbf{I}_{2}^{s}, \dots, \mathbf{I}_{K}^{s})$ of the reference subject using the visual tokenizer of the pre-trained VAR model, which capture fine-grained details of the reference subject.
Then, $\mathbf{I}^{s}$ is left-padded to the text tokens $\mathbf{C}^{t}$ and all target image token maps $\mathbf{I}$, leading to $(\mathbf{I}_{1}^s, \mathbf{I}_{2}^s, ..., \mathbf{I}_{K}^s, \mathbf{C}^{t}, \mathbf{I}_{1}, \mathbf{I}_{2}, ..., \mathbf{I}_{K})$.
Compared to the interleaved conditioning strategy $(\mathbf{C}^{t}, \mathbf{I}_{1}^s, \mathbf{I}_{1}, \mathbf{I}_{2}^s, \mathbf{I}_{2}, ..., \mathbf{I}_{K}^s, \\\mathbf{I}_{K})$ used in~\cite{li2024controlvar}, our design simplifies autoregressive dependencies and eliminates the need for joint control-image modeling.
Finally, we enforce full conditioning on $(\mathbf{I}^s, \mathbf{C}^t)$ and causal dependence on previous scales $(\mathbf{I}_{1}, \mathbf{I}_{2}, ..., \mathbf{I}_{k-1})$ for the prediction of $\mathbf{I}_k$ through attention masking.
This allows VAR model to effectively leverage multi-scale control signals from the reference subject at each scale while preserving its native next-scale prediction paradigm.
%

\subsection{Multi-Reward Reinforcement Learning}
Since both subject consistency and semantic alignment are critical for subject-driven image generation, we further incorporate Group Relative Policy Optimization (GRPO)~\cite{grpo} to fine-tune our DreamVAR, which significantly enhances the generation quality. The process is shown in Fig.~\ref{fig:var} (b).
Specifically, for a given text prompt $c^t$ and a reference subject $c^s$, a group of $G$ images ${\lbrace o_{i}\rbrace}_{i=1}^{G}$ is sampled from the old policy $\pi_{\theta_{old}}$ (i.e., the old version of our DreamVAR).
Each generated image $o_i$ is evaluated by reward functions to obtain the corresponding reward $R_{i}$. 
The advantage of $o_i$ is then calculated by normalizing its reward using statistics of the group rewards $\lbrace R_{i}\rbrace_{i=1}^{G}$:
\begin{equation}
\small
    A_{i} = \frac{R_{i} - \textrm{mean}(\lbrace R_{i}\rbrace_{i=1}^{G})}{\textrm{std}(\lbrace R_{i}\rbrace_{i=1}^{G})}.
\end{equation}
The training objective of GRPO can be formalized as:
\begin{equation}
\small
\begin{aligned}
\mathcal{J}_{\textrm{GRPO}}(\theta) =&\mathbb{E}_{c\sim C,~ \lbrace o_i \rbrace_{i=1}^{G} \sim \pi_{\theta_{old}}(\cdot|c} \Bigg\lbrack \frac{1}{{\sum_{i=1}^{G}|o_{i}|}} \sum_{i=1}^{G} \sum_{t=1}^{|o_{i}|} \\
&  \bigg( 
    \textrm{min}\big(r_{i,t}(\theta)A_{i}, \textrm{clip}(r_{i,t}(\theta), 1 -\epsilon, 1 + \epsilon)A_{i}\big)  \\ &- \beta D_{\textrm{KL}} (\pi_{\theta} | \pi_{\textrm{ref}})
    \bigg) \Bigg\rbrack,
\end{aligned} \label{grpo}
\end{equation}
where $r_{i,t}(\theta) = \frac{p_{\theta}(o_{i,t}|c)}{p_{\theta_{old}}(o_{i,t}|c)},~c=(c^t, c^s)$.
$D_{\textrm{KL}} (\pi_{\theta} | \pi_{\textrm{ref}})$ measures KL-divergence between the current policy $\pi_{\theta}$ and the reference model $\pi_{\textrm{ref}}$ (i.e., the frozen version of our DreamVAR), and $\beta$ is the weighting coefficient.
$|o_{i}|$ denotes the length of the generated image token sequence.

Two rewards are devised to jointly enhance appearance preservation and image-text alignment during GRPO training:

\noindent
\textbf{Semantic Alignment Reward ($R^{S}$):} This reward measures image-text alignment by computing the cosine similarity between the generated image and text prompt using CLIP.

\noindent
\textbf{Subject Consistency Reward ($R^{I}$):} This reward evaluates the preservation of visual details by computing cosine similarity between the subject of the generated image and reference subject in visual feature space such as~\cite{caron2021emerging,tip}.
%
Particularly, the generated and reference subjects are segmented before the reward calculation to minimize the impact of the background.

The final reward $R$ for each generated image is a weighted combination of $R^{I}$ and $R^{S}$, defined as:
\begin{equation}
\small
    R = \alpha R^{I} + \gamma R^{S},
    \label{eq:weight}
\end{equation}
where $\alpha$ and $\gamma$ are trade-off coefficients.

\subsection{Multi-Stage Training} \label{sec:threestage}
We adopt a multi-stage training strategy to progressively optimize our DreamVAR as follows:

\noindent
\textbf{Stage 1: Task Adaptation.} In the first stage, we extend the text-to-image VAR model to support subject-driven image generation by employing full fine-tuning on Subject-200K.

\noindent
\textbf{Stage 2: Supervised Fine-Tuning.} After task adaptation, we observe that the visual quality of generated images is unsatisfactory, primarily due to low image fidelity in Subject-200K.
Therefore, we curate a high-quality dataset (\textbf{DreamSubject-14K}) for supervised fine-tuning of DreamVAR.

\noindent
\textbf{Stage 3: Reinforcement Learning.} In the final stage, GRPO is applied to further optimize our DreamVAR initialized from the previous stage using subject consistency reward and semantic alignment reward. 

\section{Experiments}
\begin{figure*}
    \centering
    \includegraphics[width=0.84\textwidth]{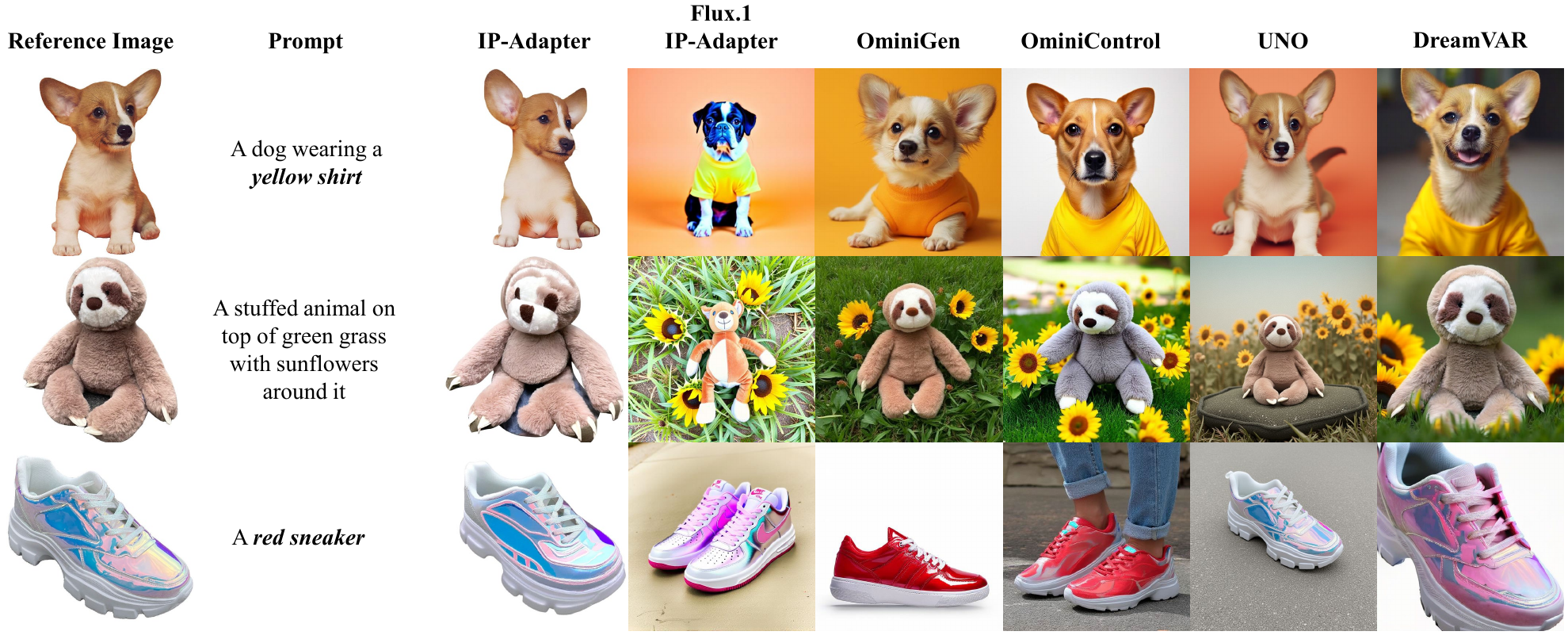}
    \vspace{-4mm}
    \caption{Qualitative comparisons with different methods on Dreambench.}
    \label{fig:compare}
    \vspace{-5mm}
\end{figure*}

\subsection{Experiments Setup}
\noindent\textbf{Datasets.}
Our DreamVAR is trained on Subject-200K~\cite{tan2024ominicontrol} to establish the initial capability of subject-driven generation.
%
To obtain DreamSubject-14K for supervised fine-tuning and reinforcement learning, we first use~\cite{yang2025qwen2} to generate diverse descriptions for each category from Object365~\cite{shao2019objects365}.
For each description, a text-to-image model~\cite{flux2024} is utilized to generate 5 images.
%
Low-quality images are filtered using~\cite{Qwen2.5-VL}.
The remaining images and the categories of the corresponding objects in the images are fed into an open-vocabulary detector~\cite{liu2024grounding} for subject localization, which are further refined by~\cite{kirillov2023segment} to produce precisely segmented subjects.
Images that fail either object detection or segmentation are discarded. 

%
\noindent
%
%
%

%
%

\noindent
\textbf{Implementation Details.}
We build our DreamVAR upon Infinity-2B~\cite{han2025infinity} by adapting it for subject-driven image generation via multi-stage training described in Sec.\ref{sec:threestage} at a resolution of $512 \times 512$.
%
In the first stage of task adaptation, the model is trained for 5,000 steps with a batch size of 16 per GPU and a learning rate of $1 \times 10^{-5}$. 
The model is fine-tuned for another 1,000 steps with a learning rate of $1 \times 10^{-4}$ in the second stage (i.e., supervised fine-tuning). 
In the final stage of reinforcement learning, the model is optimized for 3,000 steps with a batch size of 1 and a group size $G=8$ per GPU. The learning rate is $5 \times 10^{-6}$ and $\beta = 0.04$.

\noindent
\textbf{Evaluation Metrics.}
All methods are evaluated on Dreambench~\cite{ruiz2023dreambooth}.
%
Subject consistency is measured by the cosine similarity between the generated and reference subjects in DINO and CLIP feature spaces, denoted as \textbf{DINO} and \textbf{CLIP-I} scores. 
Image-text alignment is measured by the cosine similarity between the text prompt and the generated image in CLIP feature space, denoted as the \textbf{CLIP-T}.

\subsection{Experimental Results}
\noindent
\textbf{Quantitative Comparisons.}
We compare our DreamVAR against existing state-of-the-art methods in Table~\ref{tab:sota}.
As shown, despite having only 2B parameters, DreamVAR achieves comparable image-text alignment and superior subject consistency compared to the strong baselines with 12B parameters (i.e., OminiControl and UNO).
Notably, our DreamVAR yields the highest DINO (0.764) and CLIP-I (0.838) scores, surpassing all baselines.
%
These results highlight the effectiveness of DreamVAR, reinforced by GRPO, in subject-driven image synthesis, demonstrating its controllability in preserving visual details of reference subjects.

\noindent
\textbf{Qualitative Comparisons.}
As shown in Fig.~\ref{fig:compare},  DreamVAR consistently outperforms the other methods.
%
In the first and third rows, IP-Adapter, OmniGen, and OminiControl fail to preserve the appearance of the dog and sneaker, whereas UNO deviates from the prompts. In contrast, DreamVAR produces results that are more faithful to both the appearance and the prompt.
In the second row, DreamVAR successfully reconstructs the stuffed animal and aligns with the text, while the other methods fail to preserve its fine-grained details.
%

\begin{table}[]
    \centering
    \small
    \begin{tabular}{c|ccc}
    \toprule
         Method &DINO $\uparrow$ &CLIP-I $\uparrow$ &CLIP-T $\uparrow$  \\
         \hline
         DreamBooth~\cite{ruiz2023dreambooth} &0.668 &0.803 &0.305\\
         IP-Adapter~\cite{ipa}&0.696 &0.807 &0.278\\
         ELITE~\cite{wei2023elite}&0.647 &0.772 &0.296\\
         BootPIG~\cite{purushwalkam2024bootpig}&0.674 &0.797 &0.311\\
         OmniGen~\cite{xiao2025omnigen}&0.693 &0.801 &\textbf{0.315}\\
         OminiControl~\cite{tan2024ominicontrol}&0.684 &0.799 &0.312\\
         Flux.1 IP-Adapter~\cite{flux-ipa}& 0.582 &0.820 &0.288\\
         UNO*~\cite{wu2025less}&0.717 &0.819 &0.302 \\
         UNO~\cite{wu2025less}&0.747 &0.829 &0.303 \\\hline
         DreamVAR~(Ours)&\textbf{0.764}&\textbf{0.838} &0.310 \\
         \bottomrule
    \end{tabular}
    \vspace{-3mm}
    \caption{Quantitative results on Dreambench. * indicates models trained on the same data as our DreamVAR.}
    \vspace{-5mm}
    \label{tab:sota}
\end{table}

\begin{figure}
    \centering
    \includegraphics[width=0.86\linewidth]{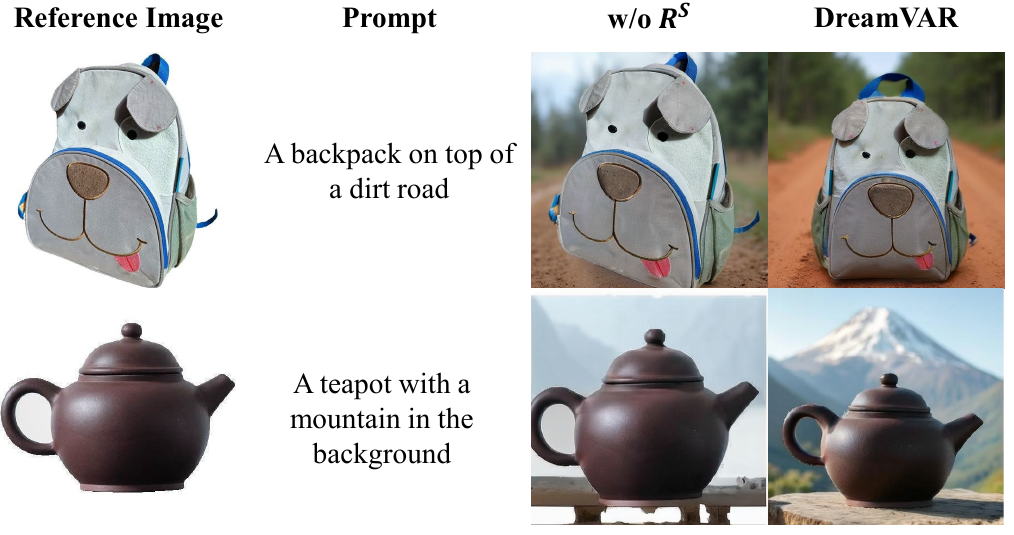}
     \vspace{-5mm}
    \caption{Ablation study of reinforcement learning rewards.}
    \label{fig:rl_ablation}
    \vspace{-3mm}
\end{figure}

\subsection{Ablation Study and Discussions}
\textbf{Effect of different rewards.} We conduct an ablation study to evaluate the effect of different reward functions. The results are listed in Table~\ref{tab:ablation}.
It can be observed that using only subject consistency reward $R^{I}$ substantially enhances detail preservation but also reduces image-text alignment.
This behavior indicates reward hacking: the model trickly inflates subject consistency reward by directly replicating the reference subject.
By additionally incorporating semantic alignment reward $R^{S}$, our full DreamVAR improves the DINO/CLIP-T score from 0.744/0.307 to 0.764/0.310.
%
Fig.~\ref{fig:rl_ablation} further compares the generated images of DreamVAR w/o $R^S$ and full DreamVAR.

\begin{table}[]
\small
    \centering
    \begin{tabular}{cc|ccc}
    \toprule
        $R^{I}$ & $R^{S}$ &DINO $\uparrow$ &CLIP-I $\uparrow$ &CLIP-T $\uparrow$ \\
        \hline
         -&- &0.744 &0.830 &0.307 \\
          \checkmark&- &0.810 &0.870 &0.270 \\
           \checkmark &\checkmark &0.764 &0.838 &0.310 \\
           \bottomrule
    \end{tabular}
    \vspace{-2.5mm}
    \caption{Ablation study of reinforcement learning rewards.}
    \label{tab:ablation}
    \vspace{-5mm}
\end{table}

\noindent
\textbf{Effect of trade-off coefficients in Eq. \ref{eq:weight}.}
We ablate the trade-off coefficients in Eq.~\ref{eq:weight} by varying their values from 1 to 3.
%
Based on the results in Table~\ref{tab:weight}, we set the default values to $\alpha = 1.0$ and $\gamma = 2.0$ to achieve the optimal balance between subject consistency and semantic alignment.

\noindent
\textbf{Effect of multi-scale subject features.}
We compare DreamVAR with multi-scale subject features versus last-scale features in Table~\ref{tab:mul-scale}.
As shown, using solely last-scale features yields lower DINO and CLIP-I scores compared to the run using multi-scale features, indicating that multi-scale features capture richer details and holistic visual context for better appearance preservation of the reference subject.

\noindent
\textbf{Effect of subject feature pre-filling.}
We also compare multi-scale feature injection using the interleaved and our pre-filling strategies in Table~\ref{tab:mul-scale}.
It shows that the ``Interleaved'' strategy lags behind both the ``Pre-filling'' method and its single-scale variant ``Last Scale''.
%
This validates that pre-filling strategy alleviates the train-test discrepancy and improves performance.

\noindent
\textbf{Computational efficiency.}
DreamVAR comprises about 2B parameter, which is more compact than UNO (12B). DreamVAR achieves a 1.75× training speedup over UNO and reduces inference time from 17 seconds to 2 seconds per image.

\begin{table}[]
\small
    \centering
    \begin{tabular}{cc|ccc}
    \toprule
        $\alpha$ & $\gamma$ &DINO $\uparrow$ &CLIP-I $\uparrow$ &CLIP-T $\uparrow$ \\
        \hline
         1.0&1.0 &0.770  &0.841  &0.304  \\
           1.0& 2.0 &0.764 &0.838 &0.310 \\
           1.0 &3.0 &0.760  &0.834  &0.308 \\
           2.0 &1.0 &0.774  &0.843  &0.300 \\
           3.0 &1.0 &0.777  &0.846 &0.300 \\
           \bottomrule
    \end{tabular}
    \vspace{-2.5mm}
    \caption{Ablation study of trade-off coefficients in Eq. \ref{eq:weight}.}
    \label{tab:weight}
    \vspace{-2mm}
\end{table}

\begin{table}[]
\small
    \centering
    \begin{tabular}{c|ccc}
    \toprule
         &DINO $\uparrow$ &CLIP-I $\uparrow$ &CLIP-T $\uparrow$ \\
        \hline
          Last Scale &0.734 &0.820 &0.306 \\
          Multiple Scales &0.744 &0.830 &0.307 \\ \hline
          Interleaved &0.710 &0.812 &0.301 \\
          Pre-filling &0.744 &0.830 &0.307 \\
           \bottomrule
    \end{tabular}
    \vspace{-2.5mm}
    \caption{Ablation study of multi-scale feature pre-filling.}
    \label{tab:mul-scale}
    \vspace{-5mm}
\end{table}

\section{Conclusion}

In this work, we investigate the potential of Visual Autoregressive (VAR) models in subject-driven image generation and further validate the efficacy of reinforcement learning in enhancing generation quality within VAR architecture.
%
Specifically, we present DreamVAR, a novel framework that builds upon a text-to-image VAR model.
To mitigate the train-test discrepancy inherent in the interleaved conditioning strategy, our DreamVAR pre-fills multi-scale subject features prior to the prediction of target image token sequence.
This design simplifies autoregressive dependencies and eliminates the need for joint control-image modeling.
%
We further incorporate Group Relative Policy Optimization to optimize our DreamVAR with subject consistency reward and semantic alignment reward.
%
%
Extensive experiments demonstrate that our DreamVAR faithfully retains subject details, outperforming state-of-the-art diffusion-based methods.

\noindent
\textbf{Acknowledgements:} This work was supported by the Key Science \& Technology Project of Anhui Province (No. 202523o09050002) and Basic Research Program of Jiangsu Province (Grant No. BK20243018).
\small
\bibliographystyle{IEEEbib}
\bibliography{strings,refs}

@inproceedings{wu2025less,
  title={Less-to-more generalization: Unlocking more controllability by in-context generation},
  author={Wu, Shaojin and Huang, Mengqi and Wu, Wenxu and Cheng, Yufeng and Ding, Fei and He, Qian},
  booktitle={ICCV},
  year={2025}
}

@inproceedings{xiao2025omnigen,
  title={Omnigen: Unified image generation},
  author={Xiao, Shitao and Wang, Yueze and Zhou, Junjie and Yuan, Huaying and Xing, Xingrun and Yan, Ruiran and Li, Chaofan and Wang, Shuting and Huang, Tiejun and Liu, Zheng},
  booktitle={CVPR},
  year={2025}
}

@inproceedings{purushwalkam2024bootpig,
  title={Bootpig: Bootstrapping zero-shot personalized image generation capabilities in pretrained diffusion models},
  author={Purushwalkam, Senthil and Gokul, Akash and Joty, Shafiq and Naik, Nikhil},
  booktitle={ECCV},
  year={2024}
}

@inproceedings{ruiz2023dreambooth,
  title={Dreambooth: Fine tuning text-to-image diffusion models for subject-driven generation},
  author={Ruiz, Nataniel and Li, Yuanzhen and Jampani, Varun and Pritch, Yael and Rubinstein, Michael and Aberman, Kfir},
  booktitle={CVPR},
  year={2023}
}

@inproceedings{wei2023elite,
  title={Elite: Encoding visual concepts into textual embeddings for customized text-to-image generation},
  author={Wei, Yuxiang and Zhang, Yabo and Ji, Zhilong and Bai, Jinfeng and Zhang, Lei and Zuo, Wangmeng},
  booktitle={ICCV},
  year={2023}
}

@inproceedings{tan2024ominicontrol,
  title={Ominicontrol: Minimal and universal control for diffusion transformer},
  author={Tan, Zhenxiong and Liu, Songhua and Yang, Xingyi and Xue, Qiaochu and Wang, Xinchao},
  booktitle={ICCV},
  year={2025}
}

@misc{flux-ipa,
    author = {InstantX Team},
    title = {InstantX FLUX.1-dev IP-Adapter Page},
    year = {2024},
}

@inproceedings{rombach2022high,
  title={High-resolution image synthesis with latent diffusion models},
  author={Rombach, Robin and Blattmann, Andreas and Lorenz, Dominik and Esser, Patrick and Ommer, Bj{\"o}rn},
  booktitle={CVPR},
  year={2022}
}

@misc{flux2024,
    author={Black Forest Labs},
    title={FLUX},
    year={2024},
    howpublished={\url{https://github.com/black-forest-labs/flux}},
}

@inproceedings{han2025infinity,
  title={Infinity: Scaling bitwise autoregressive modeling for high-resolution image synthesis},
  author={Han, Jian and Liu, Jinlai and Jiang, Yi and Yan, Bin and Zhang, Yuqi and Yuan, Zehuan and Peng, Bingyue and Liu, Xiaobing},
  booktitle={CVPR},
  year={2025}
}

@article{tian2024visual,
  title={Visual autoregressive modeling: Scalable image generation via next-scale prediction},
  author={Tian, Keyu and Jiang, Yi and Yuan, Zehuan and Peng, Bingyue and Wang, Liwei},
  journal={NeurIPS},
  year={2024}
}

@article{chung2025fine,
  title={Fine-Tuning Visual Autoregressive Models for Subject-Driven Generation},
  author={Chung, Jiwoo and Hyun, Sangeek and Kim, Hyunjun and Koh, Eunseo and Lee, MinKyu and Heo, Jae-Pil},
  journal={arXiv preprint arXiv:2504.02612}
}

@article{grpo,
  title={Deepseekmath: Pushing the limits of mathematical reasoning in open language models},
  author={Shao, Zhihong and Wang, Peiyi and Zhu, Qihao and Xu, Runxin and Song, Junxiao and Bi, Xiao and others},
  journal={arXiv preprint arXiv:2402.03300}
}

@article{ipa,
  author       = {Hu Ye and
                  Jun Zhang and
                  Sibo Liu and
                  Xiao Han and
                  Wei Yang},
  title        = {IP-Adapter: Text Compatible Image Prompt Adapter for Text-to-Image
                  Diffusion Models},
  journal      = {arXiv preprint arXiv:2308.06721}
}

@inproceedings{shao2019objects365,
  title={Objects365: A large-scale, high-quality dataset for object detection},
  author={Shao, Shuai and Li, Zeming and Zhang, Tianyuan and Peng, Chao and Yu, Gang and Zhang, Xiangyu and Li, Jing and Sun, Jian},
  booktitle={ICCV},
  year={2019}
}

@article{Qwen2.5-VL,
  title={Qwen2.5-VL Technical Report},
  author={Bai, Shuai and Chen, Keqin and Liu, Xuejing and Wang, Jialin and Ge, Wenbin and Song, Sibo and others},
  journal={arXiv preprint arXiv:2502.13923}
}

@inproceedings{kirillov2023segment,
  title={Segment anything},
  author={Kirillov, Alexander and Mintun, Eric and Ravi, Nikhila and Mao, Hanzi and Rolland, Chloe and Gustafson, Laura and Xiao, Tete and Whitehead, Spencer and Berg, Alexander C and Lo, Wan-Yen and others},
  booktitle={ICCV},
  year={2023}
}

@article{li2024controlvar,
  title={Controlvar: Exploring controllable visual autoregressive modeling},
  author={Li, Xiang and Qiu, Kai and Chen, Hao and Kuen, Jason and Lin, Zhe and Singh, Rita and Raj, Bhiksha},
  journal={arXiv preprint arXiv:2406.09750}
}

@inproceedings{ramesh2021dalle,
  title={Zero-shot text-to-image generation},
  author={Ramesh, Aditya and Pavlov, Mikhail and Goh, Gabriel and Gray, Scott and Voss, Chelsea and Radford, Alec and Chen, Mark and Sutskever, Ilya},
  booktitle={ICML},
  year={2021},
}

@inproceedings{caron2021emerging,
  title={Emerging properties in self-supervised vision transformers},
  author={Caron, Mathilde and Touvron, Hugo and Misra, Ishan and J{\'e}gou, Herv{\'e} and Mairal, Julien and Bojanowski, Piotr and Joulin, Armand},
  booktitle={ICCV},
  year={2021}
}

@article{yang2025qwen2,
  title={Qwen2.5-1m technical report},
  author={Yang, An and Yu, Bowen and Li, Chengyuan and Liu, Dayiheng and Huang, Fei and Huang, Haoyan and Jiang, Jiandong and others},
  journal={arXiv preprint arXiv:2501.15383}
}

@inproceedings{liu2024grounding,
  title={Grounding dino: Marrying dino with grounded pre-training for open-set object detection},
  author={Liu, Shilong and Zeng, Zhaoyang and Ren, Tianhe and Li, Feng and Zhang, Hao and Yang, Jie and others},
  booktitle={ECCV},
  pages={38--55},
  year={2024},
}

@inproceedings{ho2020ddpm,
	title={Denoising diffusion probabilistic models},
	author={Ho, Jonathan and Jain, Ajay and Abbeel, Pieter},
	booktitle={NeurIPS},
	year={2020}
}

@misc{openai2024gpt4technicalreport,
      title={GPT-4 Technical Report}, 
      author={OpenAI},
      year={2024},
      eprint={2303.08774},
      archivePrefix={arXiv}
}

@ARTICLE{tip,
  author={Jiang, Xin and Fang, Ziye and Shen, Fei and Gao, Junyao and Li, Zechao},
  journal={IEEE TIP}, 
  title={Progressive Feature Encoding with Background Perturbation Learning for Ultra-Fine-Grained Visual Categorization}, 
  year={2026},
}

@article{mao2025varedit,
  title={Visual autoregressive modeling for instruction-guided image editing},
  author={Mao, Qingyang and Cai, Qi and Li, Yehao and Pan, Yingwei and Cheng, Mingyue and Yao, Ting and Liu, Qi and Mei, Tao},
  journal={arXiv preprint arXiv:2508.15772},
  year={2025}
}

@article{cai2025hidream,
  title={HiDream-I1: A High-Efficient Image Generative Foundation Model with Sparse Diffusion Transformer},
  author={Cai, Qi and Chen, Jingwen and Chen, Yang and Li, Yehao and Long, Fuchen and Pan, Yingwei and Qiu, Zhaofan and Zhang, Yiheng and Gao, Fengbin and Xu, Peihan and others},
  journal={arXiv preprint arXiv:2505.22705},
  year={2025}
}

@inproceedings{wanvton,
  title={VTON-VLLM: Aligning Virtual Try-On Models with Human Preferences},
  author={Wan, Siqi and Chen, Jingwen and Cai, Qi and Pan, Yingwei and Yao, Ting and Mei, Tao},
  booktitle={NeurIPS},
  year={2025}
}

@inproceedings{chen2023controlstyle,
  title={Controlstyle: Text-driven stylized image generation using diffusion priors},
  author={Chen, Jingwen and Pan, Yingwei and Yao, Ting and Mei, Tao},
  booktitle={ACM MM},
  year={2023}
}

@article{gao2025styleshot,
  title={Styleshot: A snapshot on any style},
  author={Gao, Junyao and Sun, Yanan and Liu, Yanchen and Tang, Yinhao and Zeng, Yanhong and Qi, Ding and Chen, Kai and Zhao, Cairong},
  journal={IEEE TPAMI},
  year={2025},
}

@inproceedings{yao2025denoising,
  title={Denoising token prediction in masked autoregressive models},
  author={Yao, Ting and Li, Yehao and Pan, Yingwei and Qiu, Zhaofan and Mei, Tao},
  booktitle={ICCV},
  year={2025}
}

@inproceedings{zheng2025hierarchical,
  title={Hierarchical Masked Autoregressive Models with Low-Resolution Token Pivots},
  author={Zheng, Guangting and Li, Yehao and Pan, Yingwei and Deng, Jiajun and Yao, Ting and Zhang, Yanyong and Mei, Tao},
  booktitle={ICML},
  year={2025}
}

\end{document}